\def\year{2021}\relax
\documentclass[letterpaper]{article} 
\usepackage{aaai21}  
\usepackage{times}  
\usepackage{helvet} 
\usepackage{courier}  
\usepackage[hyphens]{url}  
\usepackage{graphicx} 

\usepackage{amssymb}
\usepackage{subfigure}
\usepackage[switch]{lineno}

\usepackage{natbib}  
\usepackage{caption} 
\title{A Self-Adjusting Fusion Representation Learning Model for \\Unaligned Text-Audio Sequences}
\author{Kaicheng Yang\textsuperscript{\rm 1,2}, Ruxuan Zhang\textsuperscript{\rm 3}, Hua Xu,\textsuperscript{\rm 1} Kai Gao\textsuperscript{\rm 2}\\ 
	\textsuperscript{\rm 1}State Key Laboratory of Intelligent Technology and Systems, Department of Computer Science and Technology, Tsinghua University, Beijing 100084, China\\
	\textsuperscript{\rm 2}School of Information Science and Engineering, Hebei University of Science and Technology,\\ Shijiazhuang 050018, China\\
	\textsuperscript{\rm 3}Northeast Electric Power University
}

\begin{document}
\maketitle

\begin{abstract}
Inter-modal interaction plays an indispensable role in multimodal sentiment analysis. Due to different modalities sequences are usually non-alignment, how to integrate relevant information of each modality to learn fusion representations has been one of the central challenges in multimodal learning. In this paper, a Self-Adjusting Fusion Representation Learning Model (SA-FRLM) is proposed to learn robust crossmodal fusion representations directly from the unaligned text and audio sequences. Different from previous works, our model not only makes full use of the interaction between different modalities but also maximizes the protection of the unimodal characteristics. Specifically, we first employ a crossmodal alignment module to project different modalities features to the same dimension. The crossmodal collaboration attention is then adopted to model the inter-modal interaction between text and audio sequences and initialize the fusion representations. After that, as the core unit of the SA-FRLM, the crossmodal adjustment transformer is proposed to protect original unimodal characteristics. It can dynamically adapt the fusion representations by using single modal streams. We evaluate our approach on the public multimodal sentiment analysis datasets CMU-MOSI and CMU-MOSEI. The experiment results show that our model has significantly improved the performance of all the metrics on the unaligned text-audio sequences.
\end{abstract}

\section{Introduction}
\begin{figure}[t]
	\centering
	\includegraphics[width=1\columnwidth]{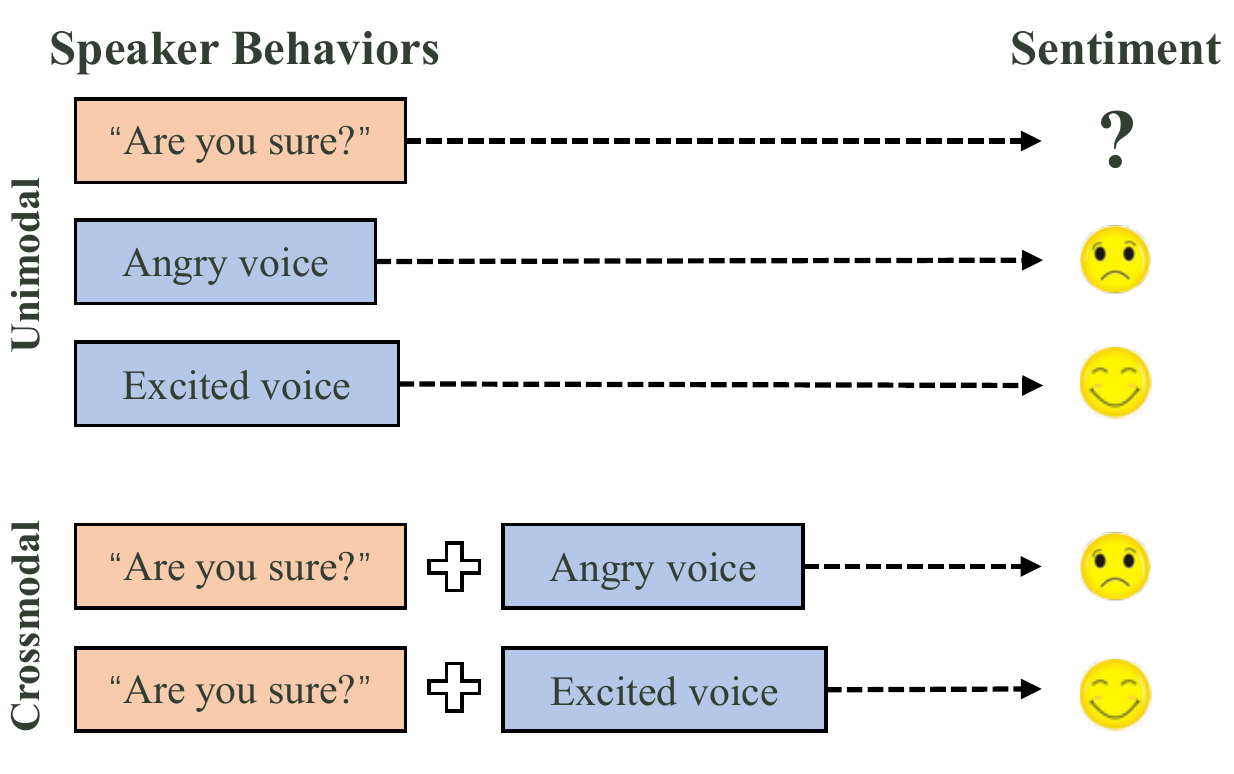} 
	\caption{Inter-modal interaction between text and audio modalities.}
	\label{fig1}
\end{figure}

People express their sentiment by using both verbal and nonverbal behaviors \cite{baltruvsaitis2018multimodal,yang2020cm}. Text as an essential modality in daily life, it expresses emotion through words, phrases, and relations \cite{turk2014multimodal}. However, the information contained in spoken words is limited. Sometimes it is not easy to identify emotions accurately only based on words. Audio is often accompanying text, and it shows sentiment by the variations in voice characteristics such as pitch, energy, vocal effort, loudness, and other frequency-related measures \cite{li2019towards,yu2020ch}. Through the inter-modal interaction between text and audio sequences, we can capture more comprehensive emotional information and improve sentiment analysis performance. Figure \ref{fig1} is an example to illustrate the inter-modal interaction between text and audio modalities. The sentiment of the sentence ''Are you sure$?$`` is ambiguous, it can express various emotions in different contexts. Nevertheless, if speak ''Are you sure$?$`` in an angry voice, it will be easily distinguished as negative. On the contrary, if ''Are you sure$?$`` is accompanied by excited voice, it will be perceived as positive.

Multimodal sentiment analysis as an increasingly extensive field of affective computing has attracted widespread attention. As one of the core problems of multimodal sentiment analysis, how to fully interact between different modalities determines the performance to a certain extent \cite{zhang2020multimodal}. Recently, there are many innovative methods have been proposed. Some researchers make the product of multimodal features as the multimodal fusion representations \cite{zadeh2017tensor}. Some other researchers perform multimodal fusion using low-rank tensors to improve efficiency, and it not only reduces the parameters but also enhances the sentiment analysis results \cite{liu2018efficient}. For better interaction, a recent attempt decomposes the multimodal fusion problem into multiple stages, and each focuses on a subset of multimodal signals \cite{liang2018multimodal}. 

However, all the approaches mentioned above are based on word-level aligned multimodal sequences and they ignore the characteristics of different modalities. In this paper, we design a Self-Adjusting Fusion Representation Learning Model (SA-FRLM) to learn robust fusion representations from unaligned text and audio sequences directly. Different from the above methods, our model not only makes full use of the interaction between different modalities but also maximizes the maintenance of the characteristics of unimodal. Specifically, we first use a crossmodal alignment module to project different modalities features to the same dimension. Then the crossmodal collaboration attention is employed to initialize the fusion representations through the inter-modal interaction between text and audio sequences. In order to maximize the preservation of the characteristics of each modality, the self-adjusting module is employed to dynamically adapt fusion representations by using single modal streams. 

To prove the effectiveness of our method, we perform a large number of experiments on the public multimodal sentiment analysis datasets CMU-MOSI \cite{zadeh2016multimodal} and CMU-MOSEI \cite{zadeh2018multimodal}. The experiments show that our model creates new state-of-the-art performance on the unaligned text and audio sequences. Compared with the baseline models, it outperforms about $1.8\%-8.4\%$ on most of the metrics. In addition, qualitative analysis proves that our model can correct sentiment intensity properly by taking into account audio modality information, and it can make more accurate predictions after adjusting the fusion representations.

The main contribution of this paper are summarized as follows:
\begin{itemize}
\item We introduce a Self-Adjusting Fusion Representation Learning Model for Text-Audio sentiment analysis, which can learn fusion representations directly from the unaligned text and audio sequences.
\item We design a novel crossmodal adjustment transformer, which can reasonably adjust fusion representations by combining single modality information.
\item We show our proposed model achieves a new state-of-the-art text-audio sentiment analysis result on the public sentiment benchmark datasets CMU-MOSI and CMU-MOSEI.
\end{itemize}

\begin{figure*}[t]
	\centering
	\includegraphics[width=1\textwidth]{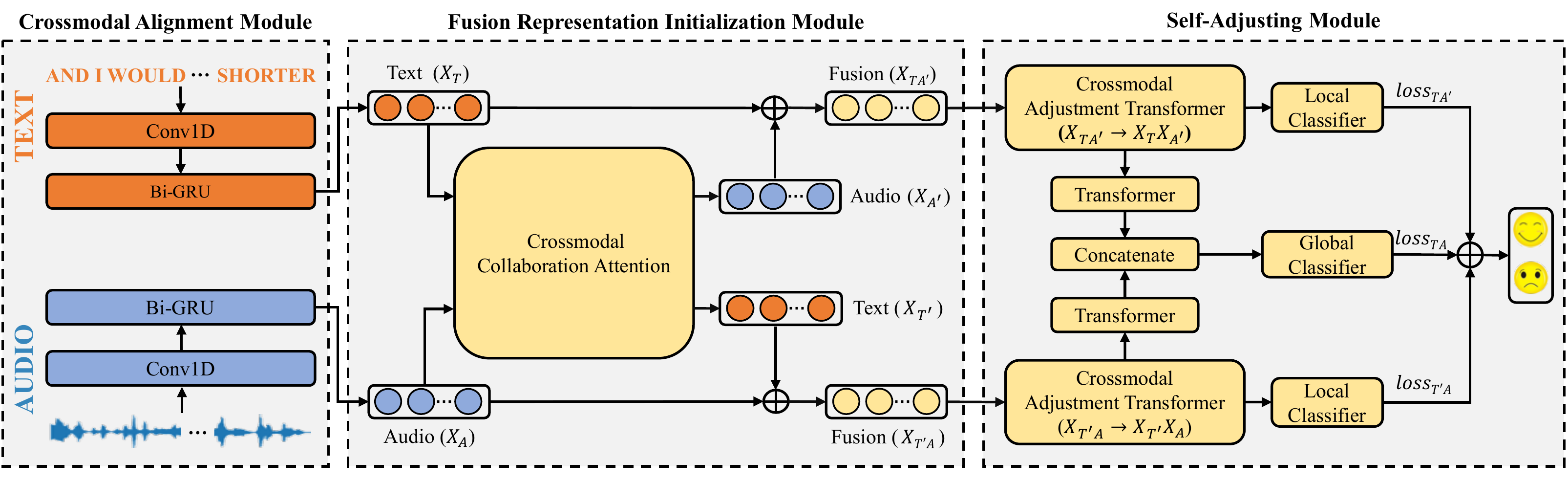} 
	\caption{Overview architecture of the Self-Adjusting Fusion Representation Learning Model.}
	\label{fig2}
\end{figure*}
\section{Related Works}

\subsection{Multimodal Sentiment Analysis}
Multimodal sentiment analysis is a new research area that aims to help machines to understand the sentiment from text, audio and video modalities \cite{xu2019multi}. Considering the internal correlation between different modalities, after fusing, we can capture more emotional relevant information and improve the performance of sentiment analysis. At present, there are many innovative models have been proposed. In the earlier, \citet{williams2018recognizing} use early fusion approach to concatenate multimodal features and get a significant improvement compared with unimodal predictors. \citet{zadeh2018multi} employ a multi-attention block and a long-short term hybrid memory to discover the interactions between different modalities. Inspired by machine translation, \citet{pham2019found} introduce a Multimodal Cyclic Translation Network (MCTN) model to learn robust joint representations by translating between modalities, it only uses text data to test and creates a new state-of-the-art result. In order to capture the dynamic nature of nonverbal intents, \citet{wang2019words} design a Recurrent Attended Variation Embedding Network (RAVEN) to dynamically shift word representations based on nonverbal cues. However, most of these methods need to align multimodal sequences on the word-level. 

With the successful use of attention mechanism in computer vision, it plays an increasingly important role in multimodal sentiment analysis. \citet{zadeh2018memory} propose a delta-memory attention network to discover both crossview and temporal interactions across different dimensions of memories in the System of LSTMs. \citet{ghosal2018contextual} introduce a Multi-modal Multi-utterance-Bi-modal Attention (MMMU-BA) framework which employs attention on multimodal representations to learn the contributing features among them. In our model, the crossmodal collaboration attention is used in the fusion representation initialization module to fully interact between text and audio modalities. Besides, we adopt crossmodal attention \cite{tsai2019multimodal} in the crossmodal adjustment transformer to latently change fusion representations by utilizing different unimodal information.

\subsection{Transformer Network}
The transformer network is first proposed for machine translation task \cite{vaswani2017attention}. Instead of recurrent neural networks and convolution neural networks, the encoder and decoder of transformer networks are based solely on attention mechanisms. Therefore, it not only has faster computation speed but also achieves better performance. In recent years, transformer networks have successfully applied in many models and frameworks. \citet{radford2018improving} adopt transformer networks in the Generative Pre-Training (GPT), which provide a more structured memory for handling long-term dependencies in text. Bidirectional Encoder Representations from Transformers (BERT) adopt bidirectional transformer networks to generate contextual word representations by jointly conditioning on both left and right context in all layers, and it has obtained new state-of-the-art results on eleven natural language processing tasks \cite{devlin2019bert}.

All the above methods only use the transformer model on text modality. How to extend it from unimodal to multimodal is still worth exploring. In the past year, \citet{tsai2019multimodal} propose the Multimodal Transformer (MulT), which is built up from multiple stacks of pairwise and bidirectional crossmodal attention blocks that directly attend to low-level features. The experiment results show that the MulT outperforms state-of-the-art methods by a large margin. However, the MulT only focuses on the interaction between modalities, and it ignores the original characteristics of different modalities. To overcome this problem, we design a crossmodal adjustment transformer which can dynamically adapt the fusion representations by importing the information of each modality.

\section{Methodology}
In this section, we introduce the architecture of our proposed Self-Adjusting Fusion Representation Learning Model (SA-FRLM). As shown in Figure \ref{fig2}, our model consists of three modules. Firstly, the crossmodal alignment module is employed to map text and audio features to the same dimension. Then these features will pass through the fusion representation initialization module to learn fusion representations through the inter-modality interaction between text and audio modalities. After that, the self-adjusting module is proposed to dynamically regulate fusion representations by using text and audio unimodal information. In the following, we first present the problem definition in Section \ref{3.1}. Then we introduce the three modules of our model in Section \ref{3.2}, Section \ref{3.3}, and Section \ref{3.4} respectively.

\subsection{Problem Definition} \label{3.1} 
Given unaligned text and audio features $F_{T}$ and $F_{A}$, firstly, we project the different modalities features to the same dimension and get the alignment text and audio representations $X_{T}$ and $X_{A}$. Then we use the attention mechanism to get the unimodal attentive representations $X_{T^{\prime}}$ and $X_{A^{\prime}}$ to initialize the fusion representations $X_{TA^{\prime}}$ and $X_{T^{\prime}A}$. The goal of this work is to improve the text-audio sentiment analysis performance by using the unimodal representations $X_{T}$, $X_{A}$, $X_{T^{\prime}}$ and $X_{A^{\prime}}$ to dynamically adapt fusion representations $X_{TA^{\prime}}$ and $X_{T^{\prime}A}$.

\subsection{Crossmodal Alignment Module} \label{3.2} 
Our proposed method aims to learn robust fusion representations directly from the unaligned text and audio sequences. After obtaining text and audio features $F_{T}$ and $F_{A}$, following \cite{tsai2019multimodal}, we pass these unaligned sequences through a 1D temporal convolutional layer. Then the different modalities sequences will be controlled to the same dimension by setting the different size of convolutional kernels and strides:
\begin{equation}
Conv_{(T, A)}=Conv1D((F_{T}, F_{A}), k_{(T, A)}, s_{(T, A)}))
\end{equation}
where $k_{(T, A)}$ and $s_{(T, A)}$ represent the number of convolutional kernels and strides for text and audio modalities. After that, two separate Bi-GRU layers are applied to extract temporal information on each modality. Finally, the aligned text and audio features $X_{T}$ and $X_{A}$ are used to initialize the fusion representations. 

\subsection{Fusion Representation Initialization Module} \label{3.3}
The purpose of the fusion representation initialization module is to initialize fusion representations through the inter-modal interaction between text and audio modalities. Inspired by MMMU-BA framework \cite{ghosal2018contextual}, we employ the crossmodal collaboration attention to enable one modality to be changed by receiving information from another modality. Specifically, we first compute a pair of attention matrices $M_{TA}$ and $M_{AT}$, which include the cross-modality information:
\begin{equation}
M_{TA}=X_{T} X_{A}^\top
\end{equation}
\begin{equation}
M_{AT}=X_{A} X_{T}^\top
\end{equation}
Then we pass the attention matrices through a $Tanh$ function and compute attention score by a $Softmax$ function, the attention score matrices $S_{TA}$ and $S_{AT}$ are defined as:
\begin{equation}
S_{TA}=Softmax(Tanh(M_{TA}))
\end{equation}
\begin{equation}
S_{AT}=Softmax(Tanh(M_{AT}))
\end{equation}
We apply soft attention to compute the modality-wise attentive representations $O_{TA}$ and $O_{AT}$:
\begin{equation}
O_{TA}=S_{TA}X_{A}
\end{equation}
\begin{equation}
O_{AT}=S_{AT}X_{T}
\end{equation}
Then the matrix multiplication is used to help different modalities focus on important information and get the attentive representations $X_{T^{\prime}}$ and $X_{A^{\prime}}$:
\begin{equation}
X_{T^{\prime}}=O_{TA} \odot X_{T}
\end{equation}
\begin{equation}
X_{A^{\prime}}=O_{AT} \odot X_{A}
\end{equation}
After that, we add $X_{T}$ and $X_{A^{\prime}}$, $X_{T^{\prime}}$ and $X_{A}$ respectively and obtain two different fusion representations $X_{TA^{\prime}}$ and $X_{T^{\prime}A}$:
\begin{equation}
X_{TA^{\prime}}=w_{T} X_{T}+w_{A^{\prime}} X_{A^{\prime}}+b_{TA^{\prime}}
\end{equation}
\begin{equation}
X_{T^{\prime}A}=w_{T^{\prime}} X_{T^{\prime}}+w_{A} X_{A}+b_{T^{\prime}A}
\end{equation}
where $w_{(T,A,T^{\prime},A^{\prime})}$ represents the weight of different unimodal representations, $b$ is the bias.

\subsection{Self-Adjusting Module} \label{3.4}
In order to keep the original characteristics of each modality, the self-adjusting module is designed to dynamically adapt the fusion representations by using unimodal streams. In the following, we elaborate on the crossmodal adjustment transformer which is the core unit of the self-adjusting module. We also introduce the self-attention transformer and classifier. 

\begin{figure}[t]
	\centering
	\includegraphics[width=1\columnwidth]{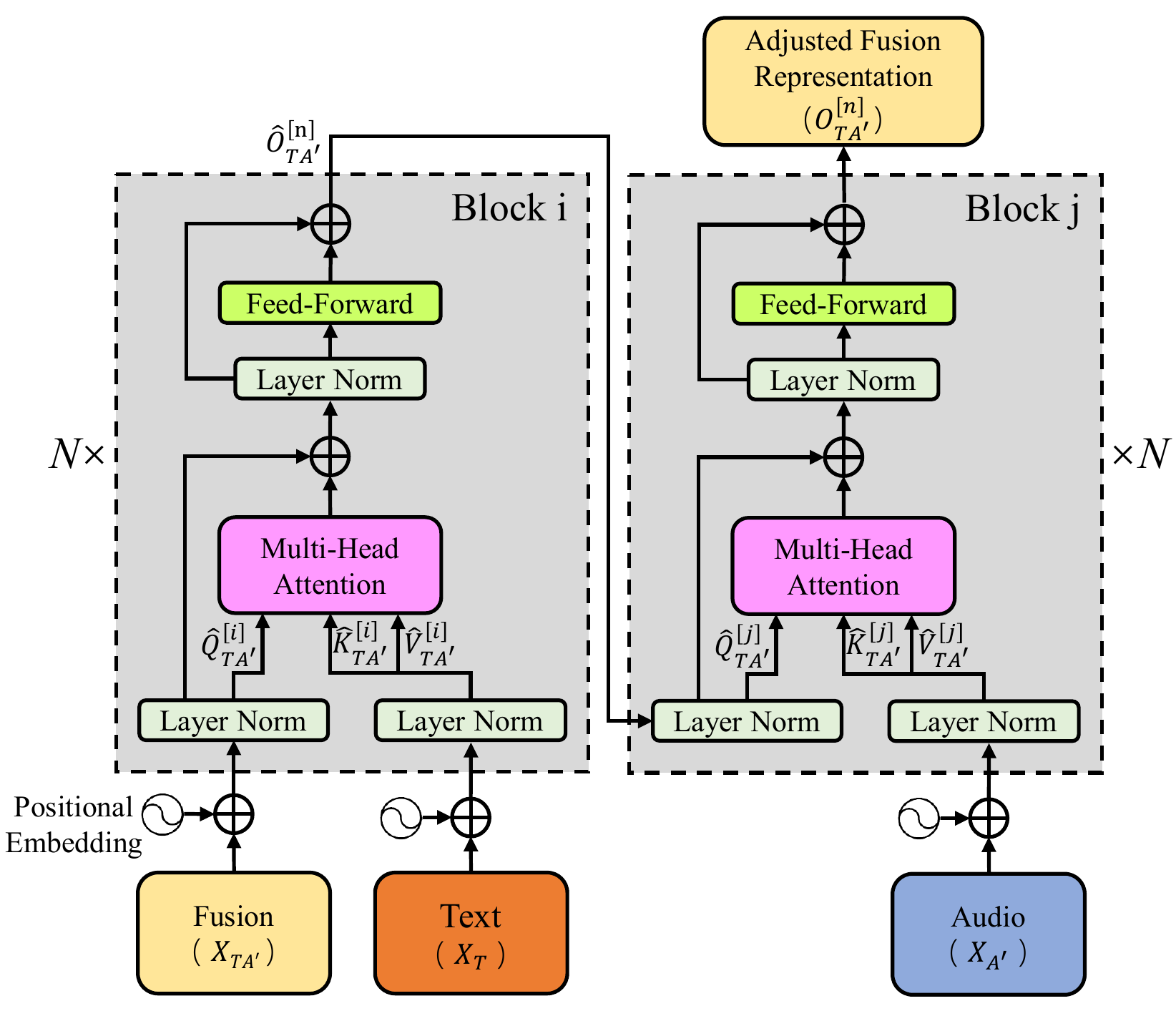} 
	\caption{The architecture of the crossmodal adjustment transformer ($X_{\mathrm{TA}^{\prime}} \longrightarrow X_{\mathrm{T}} X_{\mathrm{A}^{\prime}}$).}
	\label{fig3}
\end{figure}

\subsubsection{Crossmodal Adjustment Transformer}
Based on the previous work \cite{tsai2019multimodal}, we design the crossmodal adjustment transformer that enables fusion representations to be adjusted by utilizing different unimodal information. For convenience, we will introduce the structure of it through the example ($X_{\mathrm{TA}^{\prime}} \longrightarrow X_{\mathrm{T}} X_{\mathrm{A}^{\prime}}$), which use unimodal representations $X_{T}$ and $X_{A^{\prime}}$ to adjust the fusion representations $X_{TA^{\prime}}$.

As shown in Figure \ref{fig3}, the crossmodal adjustment transformer has three input:fusion representations ($X_{TA^{\prime}}$), text representations ($X_{T}$), and audio representations ($X_{A^{\prime}}$). To preserve the temporal information of input sequences, following \cite{vaswani2017attention} we augment position embedding ($\mathrm{PE}$) to all the input sequences. Given a sequence $X \in \mathbb{R}^{l \times d}$ ($l$ represents the sequence length and $d$ represents the feature dimension), the position embedding is computed as follows:
\begin{equation}
\mathrm{PE}_{(p o s, 2 i)}=\sin (p o s / 10000^{2 i / d})
\end{equation}
\begin{equation}
\mathrm{PE}_{(p o s, 2 i+1)}=\cos (p o s / 10000^{2 i / d})
\end{equation}
where $pos = 0,1,...,l-1$ and $i=0,1,...,\left\lfloor\frac{d}{2}\right\rfloor$. Then we add position embedding to the input sequences followed by layer normalization.
\begin{equation}
E_{\{TA^{\prime},T,A^{\prime}\}}=\mathrm{LN}(\mathrm{PE}_{\{TA^{\prime},T,A^{\prime}\}}+X_{\{TA^{\prime},T,A^{\prime}\}})
\end{equation}
where $\mathrm{LN}$ represents layer normalization. After that, we adopt $N$ crossmodal blocks to adjust the fusion representations $E_{TA^{\prime}}$ by using the text representations $E_{T}$. The crossmodal block is proposed by \cite{tsai2019multimodal}, it is mainly composed of the multi-head attention and the feed-forward layer. Besides, it employs a residual connection \cite{he2016deep} around each of two sub-layers followed by layer normalization. The Query, Key, and Value of the multi-head attention in the $i$th blocks is defined as $\hat{Q}_{TA^{\prime}}^{[i]} = \mathrm{LN}(\hat{O}_{TA^{\prime}}^{[i-1]})$, $\hat{K}_{TA^{\prime}}^{[i]}= \hat{V}_{TA^{\prime}}^{[i]} = \mathrm{LN}(E_{T})$, where $i = 1,...,N$ and $\hat{O}_{TA^{\prime}}^{[i-1]}$ is the output of the $i-1$th crossmodal block, then the multi-head attention is computed as:
\begin{equation}
\hat{Q}_{TA^{\prime}}^{[1]} = LN(E_{TA^{\prime}})
\end{equation}
\begin{equation}
MH_{TA^{\prime}}^{[i]}=Softmax(\frac{\hat{Q}_{TA^{\prime}}^{[i]} (\hat{K}_{TA^{\prime}}^{[i]})^\top}{\sqrt{d}}) \hat{V}_{TA^{\prime}}^{[i]}
\end{equation}
the output of the $i$th layers $\hat{O}_{TA^{\prime}}^{[i]}$ is defined as:
\begin{equation}
\hat{M}_{TA^{\prime}}^{[i]} = \mathrm{LN}(MH_{TA^{\prime}}^{[i]}+\hat{O}_{TA^{\prime}}^{[i-1]})
\end{equation}
\begin{equation}
\hat{O}_{TA^{\prime}}^{[i]}= \mathrm{FL}(\mathrm{LN}(\hat{M}_{TA^{\prime}}^{[i]} )+\hat{M}_{TA^{\prime}}^{[i]} 
\end{equation}
where $i = 1,...,N$ and $\mathrm{FL}$ represents the feed-forward layer.

Based on the output of the former $N$ crossmodal blocks $\hat{O}_{TA^{\prime}}^{[n]}$, we introduce another $N$ crossmodal blocks to adjust $\hat{O}_{TA^{\prime}}^{[n]}$ by using the audio feature $E_{A^{\prime}}$. The output $O_{TA^{\prime}}^{[n]}$ of the latter $N$ crossmodal blocks is the final adjusted fusion representations.

\subsubsection{Self-Attention Transformer and Classifier}
After obtaining the adjusted fusion representations $O_{TA^{\prime}}^{[n]}$ and $O_{T^{\prime}A}^{[n]}$, on the one hand, these fusion representations will be trained by the corresponding local classifier, on the other hand, the self-attention transformer is employed to capture temporal information of each fusion representation. The output of the self-attention transformer will be concatenated used to make predictions through a global classifier. More importantly, we optimize our model through a single objective function, which can make us train local classifiers and global classifier at the same time. The objective function $Loss$ is defined as:
\begin{equation}
Loss = loss_{TA^{\prime}}+loss_{T^{\prime}A}+loss_{TA}
\end{equation}

\begin{table*}[t]
	\centering
	\begin{tabular}{c|c|c|c|c|c|c}
		\hline
		Model & Modality & $Acc_{7}^{h}$ & $Acc_{2}^{h}$ & $F1^{h}$ & $MAE_{}^{l}$&$Corr_{}^{h}$ \\
		\hline
		EF-LSTM &T+A+V& 31.0 & 73.6 & 74.5 & 1.078 & 0.542 \\ 
		
		LF-LSTM  &T+A+V& 33.7 & 77.6 & 77.8 &0.988 & 0.624 \\
		
		MCTN \cite{pham2019found} &T+A+V& 32.7 & 75.9 & 76.4 & 0.991 &0.613\\
		
		RAVEN \cite{pham2019found} &T+A+V& 31.7 & 72.7 & 73.1 & 1.076 &0.544\\
		
		MulT \cite{tsai2019multimodal} &T+A+V& 39.1 & 81.1 & 81.0 & 0.889 &0.686\\
		
		\hline
		MulT \cite{tsai2019multimodal}(our run) &T+A& 34.9 & 79.2 & 79.1 & 0.991 &0.667\\
		
		SA-FRLM(ours) & T+A & \textbf{35.6} & \textbf{81.1} & \textbf{81.1} & \textbf{0.908} & \textbf{0.699}\\
		\hline
	\end{tabular}
	
	\caption{Experimental results on the CMU-MOSI dataset. $^{h}$ means higher is better and $^{l}$ means lower is better. T:text,A:audio,V:video.}
	\label{table1}
\end{table*}

\section{Experiments}
In this section, we evaluate the performance of the Self-Adjusting Fusion Representation Learning Model on the public multimodal sentiment analysis datasets CMU-MOSI and CMU-MOSEI. In the following subsections, Section \ref{4.1} shows the information about datasets and experimental settings. Section \ref{4.2} presents unimodal feature extraction. Section \ref{4.3} and Section \ref{4.4} introduce the evaluation metrics and the baseline models used in our experiments.

\subsection{Datasets and Experimental Settings} \label{4.1}
We evaluate our proposed method on the CMU Multi-modal Opinion-level Sentiment Intensity (CMU-MOSI) \cite{zadeh2016multimodal} and CMU Multimodal Opinion Sentiment and Emotion Intensity (CMU-MOSEI) \cite{zadeh2018multimodal} datasets. CMU-MOSI is composed of 93 opinion videos download from YouTube movie reviews. These videos are spanning over 2199 utterances. Each utterance is annotated in the range of [-3,+3]. The audio sequences of CMU-MOSI are extracted at a sampling rate of 12.5 Hz. Considering the speaker should not appear in both training and testing sets and the balance of the positive and negative data, we split 52, 10, 31 videos in training, validation and test set accounting for 1284, 229, and 686 utterances. Similarly, CMU-MOSEI is a multimodal sentiment and emotion analysis dataset which is made up of 23,454 movie review video clips taken from YouTube. The audio sequences of CMU-MOSEI are extracted at a sampling rate of 20 Hz. To make sure the validity of the experiment, the strategy we adopt is consistent with the previously published works \cite{tsai2019multimodal,zadeh2018multi}.

In the SA-FRLM, the number of the out channels of the temporal convolutional layer is set to 50. There are 50 units in the Bi-GRU layers, and the fully connected layers used in our model have 200 units with 0.3 dropout rate. In the training process, the number of batch size and epoch is set to 12 and 20 respectively. Besides, we use $Adam$ optimizer with 0.001 learning rate and $L1$ loss function.

\subsection{Feature Extraction} \label{4.2}
To consistent with the previous works \cite{tsai2019multimodal,rahman2020integrating}, we use the same feature extraction method for text and audio modalities.
\subsubsection{Text Feature}
We use the Glove word embeddings to embed the words sequences of video transcripts to 300 dimensional word vectors. The Glove embeddings used in our experiments are trained on 840 billion tokens from the common crawl dataset.
\subsubsection{Audio Feature}
In this work, we use the COVAREP \cite{degottex2014covarep} to extract audio features. Each segment audio file is represented as a 74 dimensional vector including 12 Mel-frequency cepstral coefficients (MFCCs), pitch and segmenting features, glottal source parameters, peak slope parameters, and maxima dispersion quotients. All of these features are extracted at a sampling rate of 100 Hz.

\subsection{Evaluation Metrics}\label{4.3}
In our experiments, consistent with previous work \cite{tsai2019multimodal}, we use the same metrics to evaluate the performance of our proposed method. 7-class accuracy ($Acc7$) is used in the sentiment score classification task, 2-class accuracy ($Acc2$) and F1 score ($F1$) are used in the binary sentiment classification task, mean absolute error ($MAE$) and the correlation ($Corr$) of model predictions with correct labels are used in the regression task. The higher value of the metrics means the better performance of the model except for $MAE$. To make experiments more convincing, we randomly select five seeds and take the average result of 5 runs as the final experimental results.
\begin{table*}[t]
	\centering
	\begin{tabular}{c|c|c|c|c|c|c}
		\hline
		Model & Modality & $Acc_{7}^{h}$ & $Acc_{2}^{h}$ & $F1^{h}$ & $MAE_{}^{l}$&$Corr_{}^{h}$ \\
		\hline
		EF-LSTM &T+A+V& 46.3 & 76.1 & 75.9 & 0.680 & 0.585 \\ 
		
		LF-LSTM  &T+A+V& 48.8 & 77.5 & 78.2 &0.624 & 0.656 \\
		
		MCTN \cite{pham2019found} &T+A+V& 48.2 & 79.3  & 79.7 & 0.631 & 0.645\\
		
		RAVEN \cite{pham2019found} &T+A+V& 45.5 & 75.4 & 75.7 & 0.664 &0.599\\
		
		MulT \cite{tsai2019multimodal} &T+A+V& 50.7 & 81.6 & 81.6 & 0.591 &0.694\\
		
		\hline
		MulT \cite{tsai2019multimodal}(our run) &T+A& 48.9 & 80.1 & 80.5 & 0.627 &0.656\\
		
		SA-FRLM(ours) &T+A & \textbf{49.9} & \textbf{80.7} & \textbf{81.2} & \textbf{0.606} & \textbf{0.673}\\
		\hline
	\end{tabular}
	\caption{Experimental results on the CMU-MOSEI dataset. $^{h}$ means higher is better and $^{l}$ means lower is better. T:text,A:audio,V:video.}
	\label{table2}
\end{table*}

\subsection{Baselines}\label{4.4}
We compare our proposed model with previous methods in multimodal sentiment analysis task. The methods we compared are as follows:

\textbf{EF-LSTM} Early Fusion LSTM (EF-LSTM) concatenates multimodal inputs and uses a single LSTM to learn the contextual information.

\textbf{LF-LSTM} Late Fusion LSTM (LF-LSTM) uses single LSTM model to learn the contextual information of each modality and concatenate the output to make predictions.

\textbf{MCTN \cite{pham2019found}} Multimodal Cyclic Translation Network (MCTN) is designed to learn robust joint representations by translating between different modalities, and it can learn joint representations using only the source
modality as input.

\textbf{RAVEN \cite{wang2019words}} Recurrent Attended Variation Embedding Network (RAVEN) models the fine-grained structure of nonverbal subword sequences and dynamically shifts word representations based on nonverbal cues, it achieves competitive performance on two publicly available datasets for multimodal sentiment analysis and emotion recognition.

\textbf{MulT \cite{tsai2019multimodal}} Multimodal Transformer (MulT) uses the directional pairwise crossmodal attention to interactions between multimodal sequences across distinct time steps and latently adapts streams from one modality to another, and it is the current state-of-the-art method on CMU-MOSI and CMU-MOSEI datasets.

\section{Results and Discussion}
\subsection{Quantitative Analysis}
In this section, we show the performance of our proposed model and compared with baseline models. In addition, we discuss the effect of the number of crossmodal blocks on experimental results.

\subsubsection{Comparison with Baseline}
The experiment results on the CMU-MOSI dataset are shown in Table \ref{table1}. Though our proposed SA-FRLM only use unaligned text and audio sequences, it significantly improved the performance on all the evaluation metrics compared with most of the baseline models using three modalities. In the binary sentiment classification task, our model achieves $81.1\%$ on $Acc_{2}^{h}$ and $F1^{h}$, which is about $3.5\%$-$8.4\%$ and $3.3\%$-$8.0\%$ improvement compared with most of the baseline models. In the sentiment score classification task, our model achieves $35.6\%$ on $Acc_{7}^{h}$, which is about 1.9 to 4.6 percentage points higher over the most baselines. In the regression task, the SA-FRLM reduces about $0.08$-$0.170$ on $MAE_{}^{l}$ and improves about $0.075$-$0.157$ on $Corr_{}^{h}$. Compared with the MulT which inputs three modalities, our model does not outperforms on $Acc_{7}^{h}$ and $MAE_{}^{l}$. This is mainly because both MulT and our proposed model are based on the transformer model. For a more fair comparison, we only input unaligned text and audio sequences to the MulT, and the experiment results show our model achieves better performances on all the evaluation metrics. Our approach improves $1.9\%$ on $Acc_{2}^{h}$, $2.0\%$ on $F1^{h}$, and $0.7\%$ on $Acc_{7}^{h}$. On $MAE_{}^{l}$ and $Corr_{}^{h}$, our model also achieves about $0.083$ and $0.032$ performance improvement.

To prove the generalization of our method to other datasets, we also perform experiments on the CMU-MOSEI dataset. The experiment results are shown in Table \ref{table2}. Similar to MOSI, our model only use text and audio sequences, and it achieves better performance compared with most of the baseline models which use three modalities. In the binary sentiment classification task, our model achieves $80.7\%$ on $Acc_{2}^{h}$ and $81.2\%$ on $F1^{h}$, which is about $1.4\%$-$5.3\%$ and $1.5\%$-$5.5\%$ improvement over most of the baseline models. In the sentiment score classification task, our model achieves $49.9\%$ on $Acc_{7}^{h}$, which improves about 1.1 to 4.4 percentage points. In the regression task, the SA-FRLM reduces about $0.018$-$0.074$ on $MAE_{}^{l}$ and improves about $0.017$-$0.088$ on $Corr_{}^{h}$. What's more, under the same experimental conditions (only input text and audio sequences), our method achieves better performance on all the metrics compared with the MulT. It improves about $0.6\%$ on $Acc_{2}^{h}$, $0.7\%$ on $F1^{h}$, and $1.0\%$ on $Acc_{7}^{h}$. On $MAE_{}^{l}$ and $Corr_{}^{h}$, it also achieves about $0.021$ and $0.017$ performance improvements.

The superior performance on the CMU-MOSI and CMU-MOSEI datasets proves the efficiency and the generalization of our proposed model. The performance of the MulT is obviously better than other baseline models, it is mainly because that it adopts the transformer instead of recurrent networks or convolutional networks. Compared with the MulT, the SA-FRLM we proposed not only fully use of the interaction between different modalities but also maximizes the protection of the unimodal characteristics. Therefore, our model significantly improves the performance on the unaligned text and audio sequences. 

\subsubsection{Effect of the Number of Crossmodal Blocks}
\begin{figure}[h]
	\centering
	\includegraphics[width=1\columnwidth]{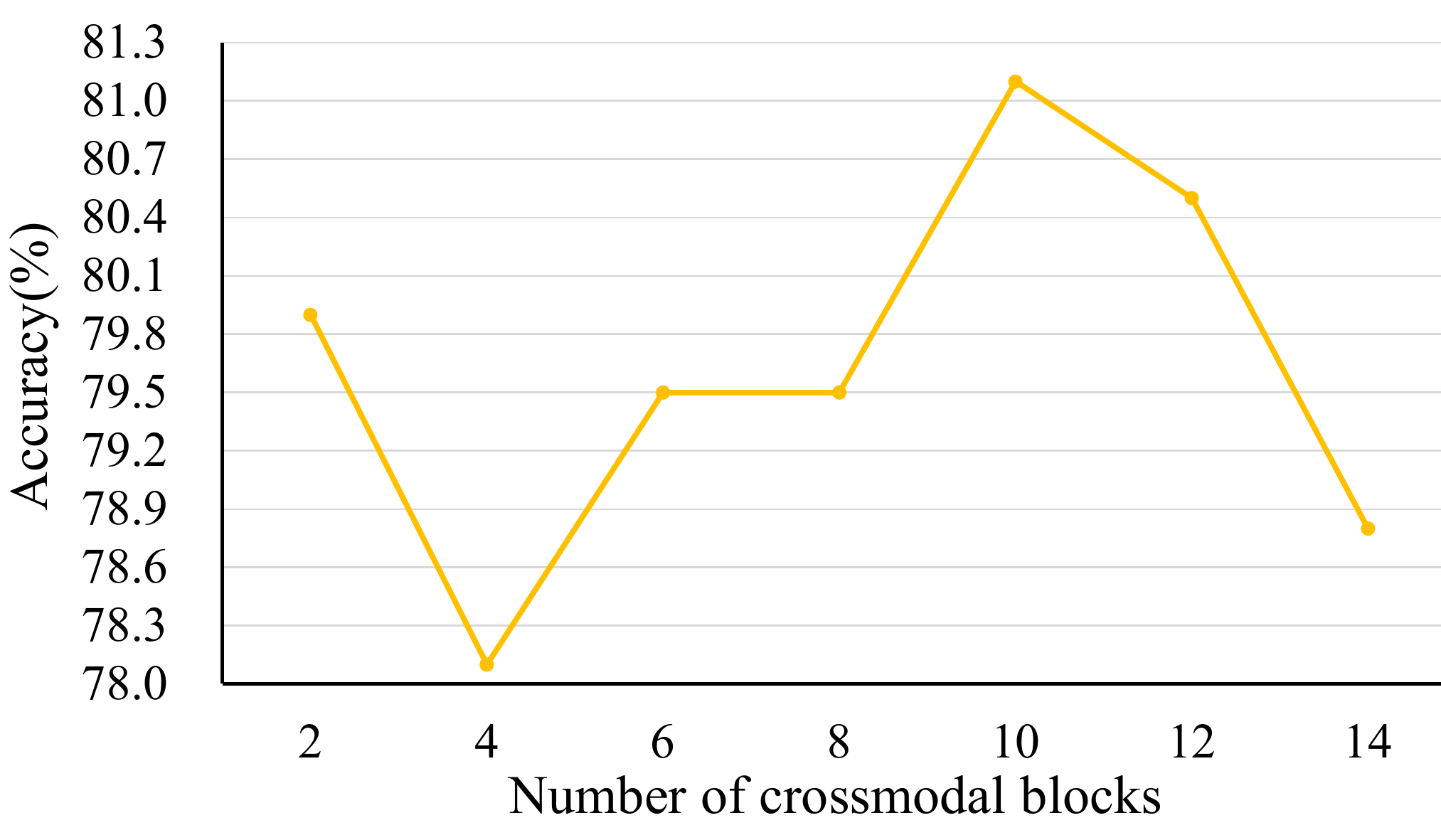} 
	\caption{The 2-class accuracy ($Acc_{2}^{h}$) of our proposed SA-FRLM with different numbers of crossmodal blocks on the MOSI dataset.}
	\label{fig4}
\end{figure}

Because the crossmodal adjustment transformer is the core unit of our model, the number of crossmodal blocks is one of the major hyper-parameters affecting performance. Figure \ref{fig4} shows the 2-class accuracy ($Acc_{2}^{h}$) of the SA-FRLM with $n$ numbers of crossmodal blocks where $n=2,4,6,...,14$ ($\frac{n}{2}$ blocks used for text modality and another $\frac{n}{2}$ used for audio modality). The results are keep growing between $4$ to $10$, and it achieves the best result with 10 crossmodal blocks. After that, with the growing number of crossmodal blocks, the complexity of our model increases gradually, which leads to a decrease of the binary accuracy.

\begin{table*}[t]
	\centering
	\includegraphics[width=1\textwidth]{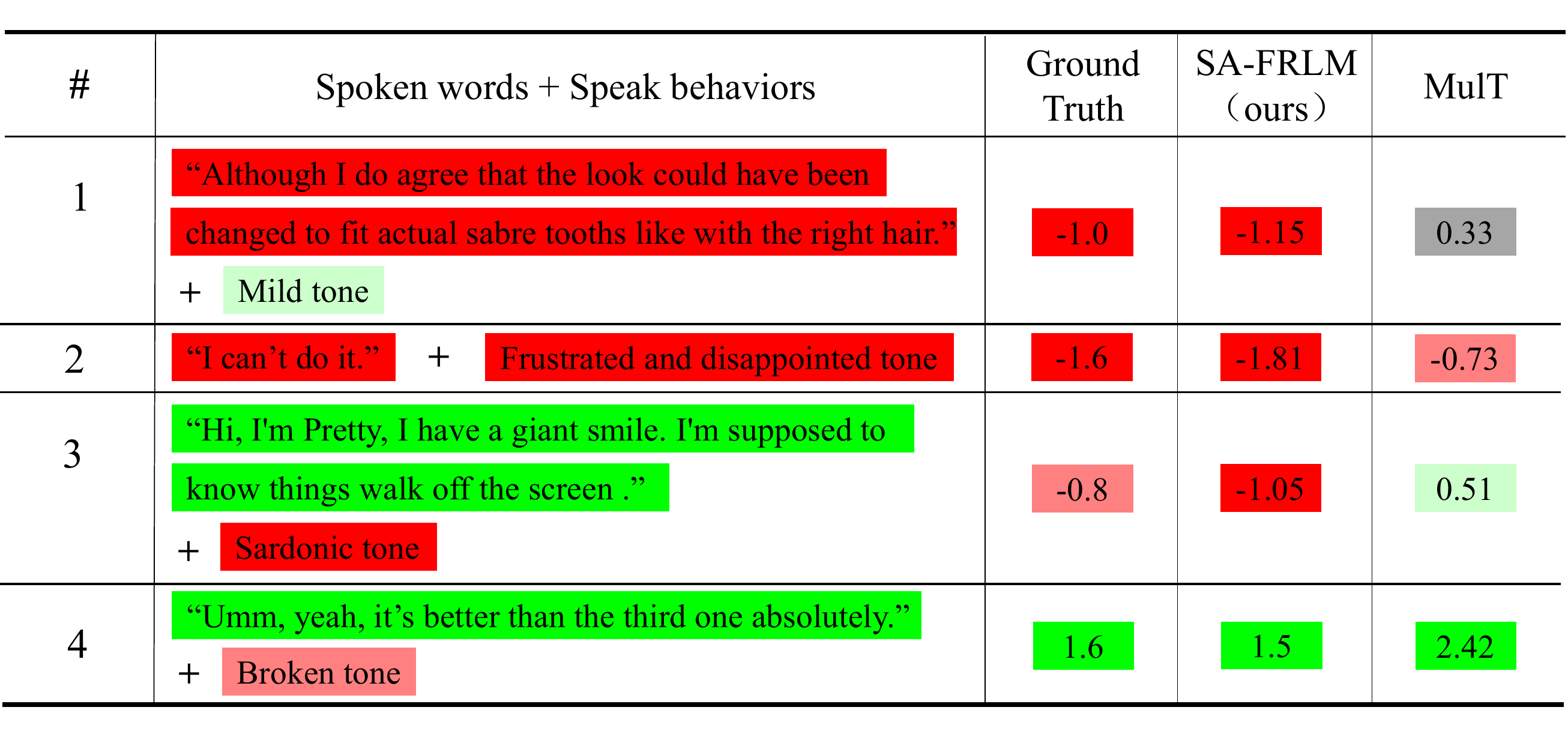} 
	\caption{Examples from the CMU-MOSI dataset. The ground truth labels are in the range of [-3,+3], where -3 represents strongly negative and +3 represents strongly positive. For each example, we show the ground truth and prediction of both the SA-FRLM and MulT.}
	\label{table3}
\end{table*}
\subsection{Qualitative Analysis}

We analyze the impact of our proposed SA-FRLM by comparing it with the MulT. As shown in Table \ref{table3}, we choose five examples from the CMU-MOSI dataset. Each example consists of spoken words as well as speak behaviors. We show the ground truth of each example and the sentiment predictions of both our model and MulT.

In the Example-1, the text has strong negative and the tone of the speaker performs weak positive. The truth label of the example is $-1.0$. Compared with the MulT, our model makes a more accurate prediction probability of $-1.15$. However, influenced by audio modality, the prediction probability of the MulT is $0.33$. In the Example-2, both text and audio information perform strong negative sentiments. Although both the SA-FRLM and MulT successfully predicted sentiment, the result of our model is more closer to the truth label. In the Example-3, the sentiment of the spoken words is strongly positive. In contrast, the audio information performs strong negative. With the help of the inter-modal interaction between text and audio modalities, our model makes a right prediction. But the MulT seems to pay more attention to text modality which causes it makes a wrong emotional judgment. This example also proves that our model can revise sentiment intensity properly by taking audio modality information into account. In the Example-4, similar to the second example, our model gets a closer prediction to the truth label compared with the MulT.

From Example-1 and Example-3, we can see that our model can better allocate the weight of different modalities. It is mainly because our approach not only makes full use of the interaction between different modalities but also maximizes the properties of the original characteristics of different modalities. In Example-2 and Example-4, compared with MulT, the prediction of our method is closer to the ground truth. The main reason is that our model dynamically adjusts the fusion representations by combining the unimodal information of different modalities. So the adjusted fusion representations are more robust and it can better represent the information of the unaligned text and audio sequences.

\section{Conclusion}
In this paper, we propose a Self-Adjusting Fusion Representation Learning Model (SA-FRLM). Different from previous works, our model not only makes full use of the interaction between different modalities but also greatly protects the characteristics of each modality. As the core unit of our model, the crossmodal adjustment transformer is proposed to dynamically change the fusion representations by combining text and audio information. The experiment results on the CMU-MOSI and CMU-MOSEI datasets show that the SA-FRLM has significantly improved the performance on the unaligned text and audio sequences. Additionally, qualitative analysis proves our model can revise sentiment intensity properly by taking audio modality information into account, and it can get more accurate predictions after adjusting fusion representations. In the future, because there are many works have proved the efficiency of the pre-trained language model, we will explore how to extend the pre-trained language model from unimodal to multimodal to improve the performance of multimodal sentiment analysis.

\bibliography{reference}
\end{document}